\pdfoutput=1
\documentclass[11pt]{article}
\ifdefined\ANON \usepackage[review]{acl} \else \usepackage{acl} \fi
\usepackage{times}
\usepackage{latexsym}
\usepackage[T1]{fontenc}
\usepackage[utf8]{inputenc}
\usepackage{microtype}
\usepackage{amsmath,amssymb}
\usepackage{booktabs}
\usepackage{graphicx}
\usepackage{multirow}
\usepackage{array}
\usepackage{subcaption}
\usepackage[ruled,vlined]{algorithm2e}
\usepackage{pgfplots}
\pgfplotsset{compat=1.18}
\usetikzlibrary{positioning,arrows.meta,shapes.geometric,fit,backgrounds}
\definecolor{accent}{HTML}{1f6f8b}
\definecolor{accent2}{HTML}{c1440e}
\definecolor{good}{HTML}{2e7d32}
\definecolor{bad}{HTML}{b71c1c}
\definecolor{soft}{HTML}{e7eef0}
\newcommand{\slaunch}{\textsc{strict-launch}}
\newcommand{\build}{\textsc{build}}
\newcommand{\ctrl}{\textsc{ctrl}}
\newcommand{\genswap}{\textsc{gen-swap}}
\newcommand{\pcand}{per-candidate}
\title{The Verifier is the Curriculum: Precision Sets the Return on Search in Code Self-Distillation}
\ifdefined\ANON
  \author{Anonymous ARR submission}
\else
\author{%
  Chenyu Zhou$^{1}$ \quad
  Qiliang Jiang$^{2}$ \quad
  Shuning Wu$^{3}$ \quad
  Xu Zhou$^{3}$\thanks{Corresponding author: \texttt{zhouxu\_nus@u.nus.edu}} \\[0.5em]
  {\normalsize $^{1}$School of Engineering, Institute of Science Tokyo, Japan}\\
  {\normalsize $^{2}$College of Control Science and Engineering, Zhejiang University, China}\\
  {\normalsize $^{3}$Department of Electrical and Computer Engineering, National University of Singapore, Singapore}\\[0.3em]
  {\footnotesize\texttt{zhou.c.76d6@m.isct.ac.jp} \quad \texttt{jiangqiliang@zju.edu.cn}}\\
  {\footnotesize\texttt{shuningwu@u.nus.edu} \quad \texttt{zhouxu\_nus@u.nus.edu}}
}
\fi
\begin{document}
\maketitle

\begin{abstract}
Post-training a code generator against a learned judge can optimize proxy
features that raise the score without improving the artifact. We study the
opposite signal: a \emph{deterministic, judge-free} filter that asks only whether a generated project launches cleanly under a
headless engine (\slaunch{}). Under this gate, rejection-sampling
self-distillation \emph{compounds} out-of-family generalization: on
GameCraft-Bench, a 14B model raises the \pcand{} clean-launch rate on four held-out families from $8.8\%$ to $\mathbf{42.2\%}$, and coverage at $32$ candidates from $84\%$ to $\mathbf{100\%}$, the gold references' own
ceiling, beating the supervised model on \textbf{every one} of the $25$ held-out tasks; retraining under three independent seeds reproduces the three-round climb at $13.3$, $26.4$, and $43.7\%$. The gate costs one engine invocation per candidate: no reward model, no judge.

At a fixed admitted count, what governs the loop is \textbf{verifier
precision}. Swapping only the gate for a lenient \build{} check
erases the gain ($p\!=\!0.0012$); a matched gold-duplication control
regresses below the supervised model ($p\!=\!0.018$). Under a \emph{semantic} gate on APPS,
dialing fuel precision $\pi$ from $1.0$ to $0.25$ at fixed candidate count prices that fuel linearly: over $23$ training seeds, half-clean fuel returns $+3.59$ percentage points (pp) against the $+3.69$ a linear rate predicts.

Under count-matched rejection-SFT only one direction of verifier error
carries a measurable cost. Masking three quarters of every problem's correct candidates moves held-out transfer by $-0.03$pp $[-0.74,+0.63]$, while admitting a quarter of false
positives moves it by $-1.58$pp. Search obeys the same gate:
quadrupling the harvest budget is worth $+1.62$pp behind a \textsc{strict} gate and nothing distinguishable from zero behind a \textsc{partial} one. Recall is nearly free; search pays
only through a precise gate: the verifier is the curriculum.
\end{abstract}

\section{Introduction}

Post-training a code-generation model requires a signal that says whether a
candidate is good. The dominant choice for open-ended generation is a
\emph{learned judge}, an LLM or multimodal model prompted to score the
output. This is convenient and correlates with human preference, but it has a
structural failure mode: a learned judge maps an output to a score through
whatever surface features drive its prediction, and if any of those features is
cheap to manipulate without improving the artifact, a model optimized against
the judge will find it. GameCraft-Bench's official judge has exactly such a feature
(\S\ref{sec:diagnosis}): an agent lifts the judge's art score to $1.8\times$ by swapping solid-color placeholders for real assets with the game's code frozen.

This paper takes the opposite stance on \emph{which signal to train against}.
Rather than chase a better learned judge, we make the training filter a
\emph{deterministic, judge-free} execution check that exposes no scalar score to
optimize, and make it the gate of an iterative self-distillation loop. A candidate passes only if the
materialized project launches cleanly under a headless engine: the engine returns exit code $0$ with no parse, load, or resource error (\slaunch{})---a property of the
artifact under a fixed engine, not an opinion.

The GameCraft-Bench task \citep{gamecraft} maps a one-paragraph brief to a
\emph{complete} Godot project from scratch (\S\ref{sec:setup}). A scene referencing a missing script, or a script with a parse error, yields
nothing that runs: an artifact must execute, not merely look plausible.

Our central finding is that under the \slaunch{} gate, rejection-sampling
self-distillation \emph{compounds}: each round's gain accrues on top of the
last at a fixed harvest budget. Starting from a supervised model that
launches cleanly on held-out families only $8.8\%$ of the time, three
rounds of ``generate on the training families, keep what launches cleanly,
retrain'' lift the \pcand{} clean-launch rate to $42.2\%$ and push coverage of the
$25$ held-out tasks, at the full $32$-candidate budget, from $84\%$ to $100\%$, the gold-reference ceiling.
The gains are significant at every round and come from the self-generated fuel
rather than from mere duplication of gold: an exactly duplicated gold control hurts. In self-distillation the
acceptance filter defines the next training distribution, so what it certifies
is what the model learns: the verifier is the curriculum.

\paragraph{Contributions.}
We introduce a deterministic, judge-free launch gate: one engine invocation, no
reward model to train, no judge to query. Under it self-distillation compounds out-of-family, beating the supervised model on \textbf{every one} of the $25$ held-out tasks; an execution-grounding readout that the gate never optimizes rises over the same rounds (\S\ref{sec:results}). We identify \textbf{verifier precision} as the causal variable under two verifier \emph{classes} (execution and semantic), and measure its conversion rate: admitted correctness converts into held-out gain
at an approximately constant rate over fuel precision $\pi\!\in\![0.25,1]$
(\S\ref{sec:mechanism},~\S\ref{sec:apps}). And we give the error accounting for
that axis. Under count-matched rejection-SFT the cost is carried almost entirely
by false positives, so a gate may spend three quarters of its recall for free
as long as the surviving pool still fills the quota. The harvest budget
$K_{\mathsf{h}}$, by contrast, pays only through a precise gate: \textsc{strict}
at $K_{\mathsf{h}}\!=\!8$ beats \textsc{partial} at $32$ on a quarter of the generation and verification (\S\ref{sec:search}). Finally, we locate that accounting in the learning \emph{interface}: swap the optimizer for GRPO at a matched Youden index ($\mathrm{TPR}-\mathrm{FPR}$) and the training signal reverses, the false-negative arm drawing \emph{more} gradient and spending it pushing away from trajectories that in fact passed (\S\ref{sec:apps}).

\section{Related work}

\paragraph{Self-training and verifier quality.}
Bootstrapping a model on its own filtered generations is well established, in STaR
\citep{zelikman2022star}, V-STaR \citep{hosseini2024vstar}, ReST and ReST-EM
\citep{gulcehre2023rest,singh2024restem}, and rejection-sampling fine-tuning for
math \citep{yuan2023scaling}, with recent on-policy, code, and consensus-gated
variants \citep{zhao2026opsd,zhang2026ssd,stein2026gates}. A concurrent line asks how the \emph{quality}
of the signal, rather than the optimizer, governs the loop: weak rubric verifiers
yield proxy gains that fail a cross-family panel \citep{mahmoud2026rubric},
systematic RLVR verifier error can collapse training
\citep{egashira2026verifiererror}, and at test time verifier precision bounds
rejection-sampling gain \citep{lu2025verification}. We establish the offline,
training-time case, holding the optimizer fixed so that a matched filter swap identifies the verifier's precision as the causal variable.

\paragraph{How much precision is enough.}
Two concurrent lines put a number on verifier reliability from the online side.
\citet{plesner2026imperfect} find RLVR within $2$pp of a clean baseline at label
noise up to $15\%$, and \citet{cai2025noisyverifier} model an unreliable verifier
as a reward channel with asymmetric error rates. But a scalar rate cannot say
which direction of error to spend a budget on: the noise in
\citet{plesner2026imperfect} is symmetric and resampled each epoch, and they name
the missing piece themselves---their precision-over-recall conclusion needs ``a
systematic controlled study varying FPR and FNR independently, including fixed
noise''. That study is what \S\ref{sec:apps} reports, offline and on policy.
On the online side the sharpest prediction is
\citeauthor{rad2026rateorfate}'s \citeyearpar{rad2026rateorfate} bandit analysis
of GRPO under noisy verification, in which the drift on incorrect reasoning modes
is governed by the Youden index alone. Our two
corrupted arms sit at that same index yet do not \emph{train} alike
(\S\ref{sec:apps}).
\citet{gureja2025verificationlimits} report the complementary regime of
model-generated tests, where relaxing an all-tests-pass criterion recovers
valuable training data that a rigid one discards.

\paragraph{Amplification versus reach.}
A distinct question is whether self-training acquires capability beyond the base
or only expresses what the base already has. On a free-verifier synthetic
domain, \citet{strozzi2026amplify} report the latter and make the distinction
decidable through a pass@$K$ crossover; \citet{qi2026segap} find a persistent gap
between self-generated and oracle supervision in closed-loop self-evolution.
Both bear on the \emph{capability} the loop ends with; ours is about
the \emph{curriculum} it trains on. What we add is the precision axis both setups hold fixed: their verifiers are exact.

\paragraph{Capacity versus verifier precision.}
Self-improvement has a capacity prerequisite \citep{song2024mindthegap}, but
there the model verifies its own outputs, coupling capacity to verifier
precision. An external, deterministic verifier decouples them: we vary precision
at fixed capacity (\S\ref{sec:mechanism}) and capacity at a fixed verifier
(\S\ref{sec:capacity}).

\paragraph{Gameable judges and verifiable game generation.}
Automatic judges and benchmarks are known to be cheatable
\citep{amodei2016concrete,zheng2025cheating,impossiblebench,thaman2026rhb,mahmoud2026rubric,helff2026gamingverifiers},
and a growing line studies code/game generation against executable checks instead
\citep{gamedevbench,jamer,opengame,vgamegym,gamegenverifier}. We use
GameCraft-Bench \citep{gamecraft} to ask how a judge-free execution filter shapes
\emph{what post-training transfers}.

\section{Setup: the GameCraft-Bench task}
\label{sec:setup}

GameCraft-Bench \citep{gamecraft} poses end-to-end game synthesis: given a
short design brief, a model must emit a complete, runnable
Godot~4 \citep{godot} project. A valid output contains, at minimum, a \texttt{project.godot}
engine configuration, one or more \texttt{.tscn} scene files, the \texttt{.gd} scripts that implement behavior, and a \texttt{demo\_outputs/} trace of
recorded input events that exercises the game. The benchmark's official metric,
\textsc{Overall}, is produced by a GPT-5.5 visual judge that renders the running game and
scores it on multiple qualitative axes.

\paragraph{Why from-scratch projects are hard.}
Correctness is \emph{conjunctive} across heterogeneous components (engine config
$\wedge$ scene graph $\wedge$ scripts $\wedge$ resources): the engine refuses to
run if a scene references a script that fails to parse, if a resource path is
malformed, or if an autoload errors at startup. A snippet that almost works still
resembles its target; a project that almost coheres does not run at all.

\paragraph{Families and the held-out split.}
The benchmark groups its $140$ tasks into $15$ game families
(platformer, strategy, roguelike, simulation, \ldots). We split the families three ways to measure \emph{cross-family} rather than in-distribution generalization: $10$ for training
($111$ gold projects), four held out, which the model never trains on
(horror, rhythm, puzzle, shooter; $25$ tasks), and one ($4$ tasks) excluded from both. The self-distillation harvest likewise draws only from training-family briefs
(Appendix~\ref{app:training}).

\section{Which signal to train against}
\label{sec:diagnosis}

We call a filter \emph{gameable} if a cheap transformation of the artifact raises
its score while leaving task-relevant executable behavior unchanged. The benchmark's visual judge is gameable: even \emph{randomly chosen} CC0 sprites lift its art score, though it scores a fixed frame consistently ($\sigma\!<\!0.01$; Appendix~\ref{app:diagnosis}).

\paragraph{A verification ladder.}
We order the available signals by how much they certify
(Appendix Fig.~\ref{fig:ladder}) and train against the strongest rung that is
still a fixed engine predicate with no tunable score. \build{}, the rung below \slaunch{}, only opens the project headless and quits; parse errors are not fatal. Read as verifiers, the rungs differ in \emph{precision}: \build{} accepts $887$ of the $888$ candidates the supervised model generates on the training families; only $73$ of those ($8.2\%$) also pass \slaunch{}.  We test the consequence of a weaker rung in \S\ref{sec:mechanism}.

\section{Method: launch-gated iterative self-distillation}
\label{sec:method}

The procedure is rejection-sampling self-distillation
\citep{zelikman2022star,gulcehre2023rest,yuan2023scaling} with one deliberate
choice: the acceptance filter is the deterministic \slaunch{} gate of
the verification ladder of \S\ref{sec:diagnosis}, applied only to training-family briefs. Matched controls isolate that signal;
Appendix Algorithm~\ref{alg:loop} states one round.

\paragraph{Rounds.}
Round $0$ produces the supervised model $M_0$ (SFT), trained on the $111$ gold
training-family projects; we write RFT-r$t$ for the model after $t$ rounds of
this rejection-filtered training. Each subsequent round generates candidates with the previous model, harvests its clean self-generated set (the \emph{fuel}), appends
it to the \emph{gold} pool, which remains the anchor, and retrains a fresh LoRA
from the base. The fuel grows every round ($67$ clean
projects from $M_0$, $123$ from $M_1$, $186$ from $M_2$).

\paragraph{Generation and training.}
Each round retrains a fresh LoRA \citep{hu2021lora} on Qwen3-14B \citep{qwen3}
and harvests $K_{\mathsf{h}}\!=\!8$ candidates per brief with vLLM \citep{kwon2023vllm}; the
full configuration is in Appendix~\ref{app:training}. For harvesting, \slaunch{} is paired with a fixed
three-file minimum, held constant in every control below.

\paragraph{Evaluation protocol.}
The primary metric is the \pcand{} clean-launch rate, with all comparisons paired
by task and tested by cluster-permutation within each task ($N\!=\!25$); coverage@$K$ (a task is covered if any of its candidates launches
cleanly) is reported with the unbiased estimator $1-\binom{n-c}{K}\!/\!\binom{n}{K}$ \citep{chen2021codex} at $K\!=\!32$ ($n$ candidates pooled per task, $c$ of them clean). We write $+a/{-}b$ for the numbers of held-out tasks an arm improves and worsens. Per-configuration budgets and coverage
estimators are in Appendix Table~\ref{tab:protocol}.

\section{Results}
\label{sec:results}

\paragraph{Compounding to the gold ceiling.}
Table~\ref{tab:main} gives the main result (per-round trajectory in Appendix Fig.~\ref{fig:compound}). Under
the \slaunch{} gate, three rounds of self-distillation lift the \pcand{} clean-launch rate on the four held-out families from $8.8\%$ to $42.2\%$. Task-level coverage at the
full $32$-candidate budget rises from $84\%$ to $100\%$, the gold-reference
ceiling, with no crossover within the sampled budget (Appendix Fig.~\ref{fig:covk}). Coverage
saturates by round~2, so the round-over-round compounding is carried by the \pcand{} rate rather than by coverage. On the four tasks the supervised model never solves within its full $32$-candidate pool, it needs $25$--$35\times$ as many draws to match a single distilled draw (Appendix~\ref{app:reach}).

\begin{table}[tb]
\centering
\footnotesize
\setlength{\tabcolsep}{1.5pt}
\caption{Results on the four held-out families ($25$ tasks). The \pcand{} clean-launch rate (over four decoding seeds) carries a Wilson $95\%$ CI describing each arm's own candidate pool; all between-arm comparisons use the paired task-level tests of \S\ref{sec:method}. The \mbox{3-seed} column reports arms retrained under three training seeds (mean\,$\pm$\,s.d.).}
\label{tab:main}
\begin{tabular*}{\columnwidth}{@{\extracolsep{\fill}}l c c c}
\toprule
model & \pcand{} \% & 3-seed & cov.@$32$ \% \\
\midrule
SFT                       & $8.8_{[7.0,10.9]}$ & ---                       & $84$ \\
RFT-r1                    & $13.6_{[11.4,16.2]}$ & $13.3\!\pm\!0.7$          & $88$ \\
RFT-r2                    & $30.9_{[27.8,34.2]}$ & $26.4\!\pm\!1.6$          & $100$ \\
\textbf{RFT-r3}           & $\mathbf{42.2}_{[38.9,45.7]}$ & $\mathbf{43.7\!\pm\!1.5}$ & $\mathbf{100}$ \\
\midrule
\ctrl{} (gold-dup)          & $5.6_{[4.2,7.4]}$ & ---                       & $84$ \\
\build{}-gated              & $8.6_{[6.9,10.8]}$ & ---                       & $84$ \\
\genswap{} & $22.4_{[19.6,25.4]}$ & ---                       & $100$ \\
\midrule
gold ceiling             & $100$                & ---                     & $100$ \\
\bottomrule
\end{tabular*}
\end{table}

\paragraph{Significance at every round.}
The improvements survive paired, task-level testing (cluster-permutation,
$N\!=\!25$): round~1 over SFT gains $+4.9$pp ($95\%$ CI $[+1.9,+8.1]$, $p\!=\!0.0017$); round~3 gains $+33.5$pp ($95\%$ CI $[+27.9,+39.6]$, $p\!<\!10^{-4}$) on every one of the $25$ tasks ($\ge\!+4/32$ clean candidates each).

\paragraph{Robustness and per-family coverage.}
 Under three independent training
seeds the round-wise rates (Table~\ref{tab:main}, \mbox{3-seed} column) are monotone in every seed and end above the single-run trajectory. Every family reaches full coverage by
round~3 (Appendix Table~\ref{tab:family}).

\paragraph{The gains are functional.}
The gains are not confined to the axis the gate
reads. An execution-grounding signal that steps the headless SceneTree shows the number of held-out tasks with a grounded clean candidate (score $\ge\!0.50$;
Appendix~\ref{app:signals}) rising monotonically,
$10\!\to\!12\!\to\!14\!\to\!15$ of $25$ (Appendix Fig.~\ref{fig:ground}). A blind
brief-compliance judge, never a training signal, counts projects that both launch
and implement their brief: $5.0\!\to\!7.4\!\to\!12.4\!\to\!14.4$ per $100$
candidates (both signals and their controls in Appendix~\ref{app:signals}).

\section{Mechanism: verifier precision is the causal variable}
\label{sec:mechanism}

Three matched controls at a fixed fuel budget locate the gain.

\paragraph{Exact duplication regresses (Appendix Fig.~\ref{fig:mech}).}
\ctrl{} adds $67$ exact duplicates of gold training projects, matched per task to
round~1's fuel. It \emph{regresses below} SFT: $5.6\%$ \pcand{} against $8.8\%$ ($+7/{-}13$, $p\!=\!0.018$). It buys no coverage:
at $K\!=\!32$ \ctrl{} sits at SFT's $84\%$
(Table~\ref{tab:main}). The $67$ copies fall on the $48$ of $111$ training tasks round~1 already covers, doubling their weight. Self-distilled round~1 at the \emph{same} budget beats \ctrl{} $+16/{-}5$, $p\!<\!10^{-4}$, in every leave-one-family-out fold ($p\!\le\!0.017$).

\paragraph{Each round's fuel is better, not just larger.}
Holding budget and coverage fixed, drawing round~1's count of fuel from a later generator (\genswap{}) isolates fuel \emph{quality}: $+8.8$pp over round~1 itself
($22.4$ against $13.6$), $+20/{-}3$, $p\!<\!10^{-4}$, nearly twice round~1's gain over SFT, so the loop compounds.

\paragraph{A matched filter swap erases the gain.}
The \build{}-gated control reruns round~1's loop with the data source fixed and only
the filter changed: \slaunch{} swapped for the lenient \build{} rung
(\S\ref{sec:diagnosis}). Because \build{} accepts nearly every candidate while
\slaunch{} accepts few, the count-matched \build{} fuel is
$81\%$ \slaunch{}-broken ($54/67$). Trained on it, the model reaches only
$8.6\%$ \pcand{} clean-launch on the held-out families: indistinguishable from
SFT ($8.8\%$, paired $p\!=\!1.0$) and far below round~1's $13.6\%$
($+5/{-}16$, $p\!=\!0.0012$). Changing only the verifier's precision---nothing about
the optimizer, the budget, or the task distribution---erases the entire
first-round gain: what transfers depends on the acceptance signal. The supervised model's candidates break exactly where the gate looks: $815$ of
$888$ fail \slaunch{}, every one on a
parse error (Appendix~\ref{app:signals}).

Volume is not the binding constraint. Cutting round~1's fuel from $67$ projects to $8$ still lifts \pcand{} clean-launch to $11.3\%$ ($+2.5$pp $[+0.4,+4.8]$ vs.\ SFT; $-2.3$pp $[-5.3,+0.3]$ vs.\ full fuel).

\paragraph{Precision buys coverage.}
Across rounds the self-distilled fuel covers $48$, then $77$, then $101$ of the
$111$ training tasks ($9\!\to\!10\!\to\!10$ families) with almost no duplication,
whereas the exact-duplication control re-covers the same $48$ tasks and regresses
(Appendix Fig.~\ref{fig:mech}). High precision thus buys coverage rather than costing it, inverting the usual
trade-off \citep{kynkaanniemi2019precision}: a binary launch gate certifies only that the project runs, so rounds harvest novel working variants. The same widening appears at 8B, where the fuel reaches $31\!\to\!51\!\to\!75$ of the $111$ training tasks (Appendix~\ref{app:training}).

\section{Pricing the fuel: a semantic gate in competitive programming}
\label{sec:apps}

The precision axis above is established with an \emph{execution} gate. We test it under a second \emph{class} of verifier, one that checks correctness rather
than execution: APPS introductory competitive-programming problems
\citep{hendrycks2021apps}, where a candidate is accepted only if it passes
\emph{all} hidden unit tests. Unlike the binary launch signal, a unit-test suite admits a \emph{graded} verdict, so we can ask not only whether a precise gate beats a lenient one
but whether the \emph{most} precise gate is best.

\paragraph{Setup.}
We lightly supervise Qwen3-4B on $300$ APPS-introductory solutions (one epoch),
giving an APPS model, again $M_0$, that solves $20.4\%$ of $470$ held-out problems at pass@$1$. From
$M_0$ we harvest $K_{\mathsf{h}}\!=\!8$ candidates on each of $1500$ training problems and score every candidate on the unit-test ladder; $527$ problems yield an all-pass candidate. We then build
three \emph{count-matched} fuel sets ($1300$ candidates each, gold held
fixed) differing only in the gate that admitted them: \textsc{strict} (all tests
pass), \textsc{partial} (at least half the tests pass, but not all), and \textsc{lenient} (syntax-valid, but not all tests pass). We retrain $M_0$ on each set under the same recipe; we evaluate pass@$1$ and coverage@$8$ over eight training seeds on the $470$ held-out problems (three decoding seeds each), paired per problem.

\begin{figure}[tb]
\centering
\includegraphics[width=\columnwidth]{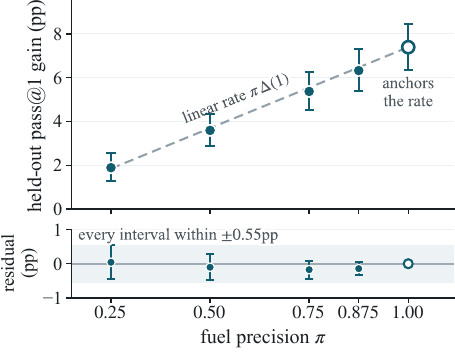}
\caption{\textbf{Linear conversion of clean fuel into transfer.} Held-out gain
over $M_0$ (pp) against fuel precision $\pi$, the fraction of the
admitted candidates that pass all tests. The APPS arms are \emph{controlled}: $\pi$ is set by construction at fixed task set, gold pool, and candidate count ($1300$). Dashed: the linear rate $\pi\,\Delta(1)$, where $\Delta(1)$ is the gain from clean fuel.
Lower panel: residuals from that rate. Bars are $95\%$ crossed seed$\times$task bootstrap CIs over $23$
training seeds and $470$ problems.}
\label{fig:dose}
\end{figure}

\paragraph{Transfer rises sharply with gate strictness (Appendix Table~\ref{tab:apps}).}
\textsc{strict}-gated self-distillation raises held-out pass@$1$ from $20.4\%$ to $25.5\%$
($+5.15$pp $[+4.14,+6.20]$, $p\!=\!0.0002$) and coverage@$8$ from $41.5\%$ to $44.8\%$ ($+3.32$pp $[+1.61,+5.06]$, $p\!=\!0.0002$) on the same $470$ problems as Fig.~\ref{fig:dose}. The controls
separate by gate strictness rather than by volume: \textsc{lenient} leaves pass@$1$ untouched ($-0.10$pp) while cutting coverage@$8$ by $3.16$pp ($p\!=\!0.0004$); \textsc{partial} moves pass@$1$ by $+2.11$pp
($p\!=\!0.0002$); and \textsc{strict} beats \textsc{partial} in a direct paired test
($+3.04$pp, $p\!=\!0.0002$). At equal candidate count this repeats the
conclusion of Godot's filter swap (\S\ref{sec:mechanism}), now under a semantic verifier
and against a graded ladder. The ordering is stable across
eight independent training seeds that vary LoRA initialization and data order:
\textsc{strict} beats every other arm in every training seed. At a fixed task mix the ordering
holds: on the $173$ \emph{training} problems that all three gates cover, \textsc{strict} still
beats \textsc{partial} ($+3.3$pp, $p\!=\!0.016$); on the $96$ \emph{held-out} problems
that $M_0$ never solves, \textsc{strict} gains $+3.4$pp ($p\!=\!0.0002$, cross-fitted on held-out decoding seeds).
The ordering does not hinge on the controls excluding all-pass candidates: against a gate
rebuilt to admit any candidate passing at least half the tests, all-pass included,
\textsc{strict} still leads by $+0.98$pp $[+0.49,+1.46]$ ($p\!=\!0.0004$) in all eight seeds.

\paragraph{A linear conversion rate.}
The ladder above varies which rule admits a candidate. To isolate precision itself with a \emph{constructed} clean arm (frozen per-problem histogram; not Appendix Table~\ref{tab:apps}'s harvested \textsc{strict}), we hold the task set, gold pool, and candidate count
fixed and vary only $\pi$, the fraction of the $1300$ fuel candidates that
actually pass all tests. Injected false positives displace clean candidates on the problems
the clean arm already covers. We measure this ladder over $23$ training seeds and $470$ held-out problems ($150$ as elsewhere,
plus $320$ further APPS problems), under the crossed seed$\times$task bootstrap
defined in Appendix~\ref{app:training}. Transfer then tracks precision at a constant rate
(Fig.~\ref{fig:dose}): the arms gain $+7.39$, $+6.32$, $+5.37$, $+3.59$, and
$+1.89$pp over $M_0$ at $\pi\!=\!1.0$, $0.875$, $0.75$, $0.5$, and $0.25$,
monotone across all five. Measured against the linear rate $\pi\,\Delta(1)$, the residuals at $\pi\!=\!0.875$, $0.75$, $0.5$, and $0.25$ are $-0.14$, $-0.17$, $-0.10$, and $+0.04$pp, each with its own $95\%$ CI covering zero and contained in $\pm0.55$pp: halving the admitted correctness returns $+3.59$pp where
the rate predicts $+3.69$. Precision therefore \emph{prices} the fuel it admits:
a batch of precision $\pi$ buys about $\pi$ of what clean fuel buys. The price is
steep. Every rung sits below clean fuel: the gap $\Delta(1)-\Delta(\pi)$ has its interval clear of zero,
from $1.07$pp $[0.84,1.30]$ at $\pi\!=\!0.875$ up to $5.50$pp $[4.76,6.29]$ at
$\pi\!=\!0.25$.

The rate also holds out of sample. The $320$ problems are disjoint from the
$150$ used elsewhere and entered only after the $150$-problem analysis was
fixed, so they are a held-out replication rather than a larger sample. On that
increment alone the five arms gain $+7.77$, $+6.66$, $+5.63$, $+3.82$, and
$+2.05$pp, strictly monotone with every interval clear of zero, and the four
residual intervals again all contain zero, within $\pm0.72$pp.
The rate is not confined to constructed ladders. Two gates built by thresholding the
unit-test fraction, admitting at $\ge\!25\%$ and at $\ge\!50\%$ with all-pass included,
realize $\pi\!=\!0.44$ and $0.58$; over eight training seeds their gains land on
$\pi\,\Delta(1)$ within $0.10$pp.

\paragraph{Only false positives are paid for (Appendix Fig.~\ref{fig:errdir}).}
The ladder above moves precision by injecting false positives; a gate can also err the other way. From a frozen
$K_{\mathsf{h}}\!=\!48$ harvest we build a second
ladder that fixes the admitted count and varies the \emph{direction} of the
error: \textsc{clean} (all-pass), \textsc{fp25} (a quarter of the slots replaced
by same-problem, syntax-valid, non-all-pass completions),
\textsc{fn25}/\textsc{fn50}/\textsc{fn75} (one quarter, one half, and three quarters of every problem's true positives masked), and \textsc{mix} (a $25\%$ error budget split evenly between the two directions). All six arms carry an identical per-problem slot histogram at a
common count-matched quota of $944$ rows, over eight training seeds, so they differ
only in which kind of error the verifier makes.

Removing three quarters of every problem's true positives moves the
held-out rate by $-0.03$pp ($95\%$ CI $[-0.74,+0.63]$), against $-1.58$pp
$[-2.23,-0.95]$ for a quarter of false positives; no false-negative rung is
distinguishable from \textsc{clean}, while every false-negative rung beats \textsc{fp25} directly ($+1.43$, $+1.74$, $+1.55$pp; every interval clear of zero).

The \textsc{mix} arm makes the accounting explicit. Splitting the error budget evenly moves the rate by $-0.87$pp $[-1.56,-0.18]$, while its false-positive half alone predicts $-0.79$pp: the two agree within $0.08$pp. At a fixed
admitted count, then, the damage is almost entirely a function of the
false-positive rate, and a gate's recall is very nearly free, because it costs no \emph{problem} coverage: the false-negative arms keep each
problem's slot count by construction. A larger harvest budget buys the
other thing: problems the loop did not cover before (\S\ref{sec:search}).

Precision is the active variable, not a stand-in for how much correct data
clears the gate: at a fixed admitted count \emph{and} a fixed number of updates,
replacing a quarter of the true positives with copies of the survivors costs
nothing measurable, while replacing them with false positives costs $2.57$pp
$[1.75,3.50]$ in all eight seeds (Appendix~\ref{app:training}).

\paragraph{The ordering belongs to the learning interface, not to the data
(Appendix Table~\ref{tab:grpo}).}
That ranking is a statement about \emph{offline} rejection sampling, where the
gate decides which rows enter a fixed-size training set: a false positive writes
a wrong trajectory into the fuel, while a false negative merely withholds one of
many interchangeable right ones. An on-policy learner meets the same two errors
on different terms, because the verifier is the reward rather than the filter: a
false negative there does not withhold a row but penalizes a trajectory the model
in fact got right. We therefore repeat the contrast with the optimizer swapped and everything else held fixed: the same two
corruption directions at a $25\%$ dose of each, but GRPO
\citep{shao2024deepseekmath} in place of rejection-SFT, over eight training seeds
per arm as in the offline experiment (run configuration in
Appendix~\ref{app:training}). Each dose is a confusion-matrix rate; because each arm perturbs one direction only, the two match the Youden index $J\!=\!\mathrm{TPR}-\mathrm{FPR}$ to within $10^{-4}$, as in \citet{rad2026rateorfate}.

Held at that index, the two directions do not train alike. A clean verifier
leaves $48.8\%$ of rollout groups with no learning signal at all: every member
earns the same reward, so the group-relative advantage vanishes. Masking a
quarter of the true positives cuts that share to $14.1\%$, and false positives cut it to $36.7\%$. Both hold in all eight training seeds. The false-negative arm
therefore receives \emph{more} gradient than the clean one in every seed,
not less (Appendix Table~\ref{tab:grpo}), and spends it pushing away from trajectories that in fact
passed: measured by the uncorrupted verifier, the policy's own rollouts score $0.608$ against the clean arm's $0.696$. Injected false positives push toward trajectories that in fact failed, and push less, because a group in which every rollout already
passes offers nothing to flip. What eight seeds resolve here is the training
signal, which separates in every one of them; held-out transfer points the same way
($\textsc{fn}\!-\!\textsc{fp}\!=\!-1.90$pp $[-5.08,+0.96]$ over all $150$ problems).

\paragraph{The loop compounds under a semantic gate too.}
A second \textsc{strict}-gated round continues the climb: harvesting from round~1 and
retraining from the base adds $+2.6$pp over round~1 ($95\%$ CI $[+1.4,+3.9]$,
$p\!<\!10^{-4}$; $+7.9$pp over $M_0$, on the $150$-problem set), with coverage@$8$ rising monotonically
($42.9\!\to\!48.0\!\to\!49.3\%$). The fuel compounds
too: the number of \emph{training} problems yielding all-pass fuel grows from
$527$ to $660$ ($+25\%$), outpacing the gain in pass@$1$. The gate widens what the loop
covers at a fixed harvest budget rather than polishing the part it already covers (Appendix~\ref{app:training}).

\section{How far this goes: scale}
\label{sec:capacity}

The \build{} swap and the APPS precision ladder both hold the base fixed and vary the verifier.
The complementary swap, varying the base at a fixed verifier, asks how far the
loop travels. We rerun the full loop (supervised fine-tuning on the same $111$ gold projects, then \slaunch{}-gated self-distillation) across model scale; across lineages we hold that recipe fixed---verifier and gold task families unchanged---and ask only whether the fuel itself is Qwen-specific.

\paragraph{Cross-scale: three rounds compound at 8B.}
At 8B (Qwen3-8B$+$LoRA, identical recipe, five decoding seeds) the loop runs the
full three rounds and compounds monotonically: \pcand{} clean-launch climbs
$3.5\!\to\!6.2\!\to\!9.3\!\to\!17.0\%$, and coverage@$32$ (unbiased, same estimator as Table~\ref{tab:main}) rises $50.8\!\to\!69.1\!\to\!83.6\!\to\!99.5\%$. The \pcand{} rate gains $+13.5$pp over the supervised model
(bootstrap $[+9.9,+17.7]$, $p\!<\!10^{-4}$), with the third round alone worth
$+7.7$pp ($[+4.3,+11.3]$, $p\!=\!0.0004$). One round buys $+2.7$pp here against
$+4.9$pp at 14B, so $1.75\times$ the parameters is worth $2.2$pp of one-round gain,
under the same recipe, gate, and task families (per-scale evaluation budgets in Appendix~\ref{app:training}). The gate's advantage persists at this scale: three launch-gated rounds stand $+12.3$pp above one \build{}-gated round
($[+8.3,+16.6]$, $p\!<\!10^{-4}$). The loop survives a halving of scale: at 4B one
launch-gated round lifts \pcand{} clean-launch from $2.00\%$ to $4.17\%$
($+2.17$pp, bootstrap $[+0.67,+3.75]$, $p\!=\!0.016$ over six decoding seeds, and
positive in all six) and coverage@$48$ from $56.0\%$ to $68.0\%$.

\paragraph{Every base produces fuel once the recipe is controlled.}
Every base we tried, across three lineages, produces \slaunch{}-clean fuel at $2.1$--$9.6\%$ per candidate on the canonical $111$ training tasks ($888$ candidates per base, $8$ per task), under a recipe matched across bases in supervised data, sequence budget, and adapter placement (Appendix Table~\ref{tab:fivebase}); the highest fuel rate in the sweep
belongs to Gemma-3-12B, above Qwen3-14B.

\section{Search is a multiplier, not a substitute}
\label{sec:search}

\begin{figure}[t]
\centering
\includegraphics[width=\columnwidth]{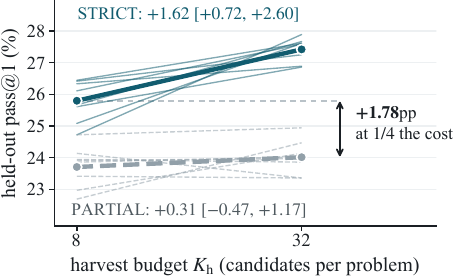}
\caption{\textbf{Search is a multiplier, not a substitute.} Held-out pass@$1$ for
the four count-matched APPS arms, as means over eight training seeds on $150$
held-out problems; thin lines are the individual seeds. Every contrast is paired
by seed.}
\label{fig:price}
\end{figure}

Back on APPS, raising only the candidates drawn
per problem to $K_{\mathsf{h}}\!=\!32$, with model, adapter, decoding, fuel
selection, training, and evaluation held fixed, gains $+2.44$pp ($[+0.97,+4.03]$, $p\!=\!0.0025$).
That invites the obvious economy: sample more and check less. It does not work.
We cross the two knobs directly on the same frozen APPS pool (Fig.~\ref{fig:price}): gate in $\{\textsc{strict},\textsc{partial}\}$ crossed with $K_{\mathsf{h}}$ in $\{8,32\}$, four count-matched arms at a common $900$-row quota, eight training seeds, all built from prefixes of the same candidates, one round from $M_0$ (a different estimand from the $+2.44$pp above).
Quadrupling the harvest is worth
$+1.62$pp $[+0.72,+2.60]$ of held-out pass@$1$ behind the \textsc{strict} gate, in all eight
seeds, and is
indistinguishable from zero behind \textsc{partial} ($+0.31$pp $[-0.47,+1.17]$); the paired \textsc{strict}-minus-\textsc{partial} premium is $+1.31$pp $[+0.85,+1.83]$ (all eight seeds positive). The \textsc{strict} gate at the \emph{low} budget still beats the \textsc{partial}
gate at four times the budget, by $+1.78$pp $[+0.65,+2.94]$ ($p\!=\!0.0026$,
in all eight seeds). And the gate's own advantage widens as the budget grows,
from $+2.09$pp $[+1.05,+3.15]$ at $K_{\mathsf{h}}\!=\!8$ to $+3.40$pp
$[+2.17,+4.67]$ at $K_{\mathsf{h}}\!=\!32$: what a larger budget returns is set by the gate that reads what it buys.

\paragraph{The three axes, priced in one currency.}
We price the axes in what the loop spends: verifier invocations and generated tokens. Sharpening the gate
is free in that currency, since \textsc{strict} and \textsc{partial} each cost one unit-test run, and \slaunch{} and \build{} each cost one engine invocation; quadrupling the
harvest costs $4\times$ the generation and $4\times$ the verification; going from 8B to 14B costs $1.75\times$ as much compute per token. The one cost precision does carry is
yield: at a fixed admitted count a stricter gate needs $1/\text{yield}$ times the
generation, and \slaunch{} admits $73$ of $888$ candidates where \build{} admits
$887$. That cost falls as the loop runs, since the generator's own clean rate
rises $4.6\!\to\!9.8\!\to\!17.5\%$ across rounds at 8B
(Appendix~\ref{app:training}). Within each domain the returns then run inversely to price. On APPS the free axis returns the $+2.09$pp above against the
$4\times$ axis's $+1.62$pp. On Godot the free axis returns $5.0$pp of
one-round gain against $2.2$pp for the $1.75\times$ axis
(\S\ref{sec:mechanism},~\S\ref{sec:capacity}). In our setting, precision is the
cheapest of the three axes and the one that returns the most.

\section{Conclusion}

The verifier is the curriculum. Under a
deterministic, judge-free launch gate, rejection-sampling self-distillation
\emph{compounds} to the gold-reference coverage ceiling on held-out families. Precision sets how high
the loop climbs, at a rate one can measure: a lenient filter erases the gain (\S\ref{sec:mechanism}), and under a semantic gate
admitted correctness converts into held-out gain at an approximately constant
rate, so fuel of precision $\pi$ retains about $\pi$ of what clean fuel buys
(\S\ref{sec:apps}).

Every base we tried, across three lineages, supplies fuel (\S\ref{sec:capacity}); on the four tasks the supervised model never solves within budget, the distilled advantage is a $25$--$35\times$ sampling-efficiency gain (Appendix~\ref{app:reach}). The gate needs no domain-specific retuning: the same all-or-nothing rule applies to a launch check and to a unit-test check. Where a verifiable predicate exists for the artifact, precision is what to buy first.

\section*{Limitations}

Our main compounding loop is studied on a single engine ($N\!=\!25$ held-out tasks);
the precision axis replicates in a second domain under a semantic
gate (APPS, \S\ref{sec:apps}). At $N\!=\!25$, only single-round differences well above
the per-task variability are resolvable, and because that variability is between tasks,
drawing more candidates per task does not reduce it; the APPS comparisons, at
$150$--$470$ problems, are not so constrained. We therefore report Godot effects as rates with intervals, always with the evaluation budget at which each clean-launch rate was measured. Execution grounding is a lower bound on functionality, covering runtime survival and
observable state change rather than brief compliance, which we measure
separately with a blind judge.

\section*{Ethical Considerations}

This work uses public code- and game-generation benchmarks (GameCraft-Bench,
APPS) and publicly available models, involves no human-subjects data, and
relies on deterministic execution checks rather than human annotation. Compute is
reported in Appendix~\ref{app:training}; the artifacts we will release use only resources whose licenses permit research use and redistribution.

\bibliography{refs}

\appendix
\setcounter{topnumber}{3}
\setcounter{bottomnumber}{2}
\setcounter{totalnumber}{5}
\renewcommand{\topfraction}{0.92}
\renewcommand{\bottomfraction}{0.6}
\renewcommand{\textfraction}{0.07}
\renewcommand{\floatpagefraction}{0.75}

\begin{figure}[tb]
\centering
\includegraphics[width=\columnwidth]{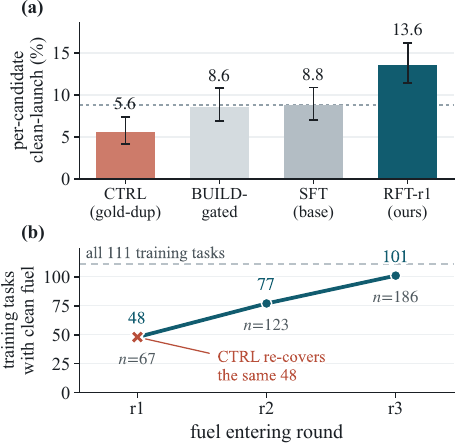}
\caption{\textbf{(a)} Same $67$-sample budget; the dashed line is the supervised model's rate. Neither control improves on it: exact gold duplication regresses ($p\!=\!0.018$) and the lenient \build{} filter merely matches it. Only \slaunch{}-filtered self-distillation gains ($p\!<\!10^{-4}$ against \ctrl{}, $p\!=\!0.0012$ against \build{}-gated). \textbf{(b)} What the precise gate buys: the training tasks, of $111$, whose fuel contains a clean project, with each round's fuel size beside its point. The exact-duplication control re-covers the $48$ that round~1's fuel already reaches.}
\label{fig:mech}
\end{figure}

\section{Diagnosis: the visual judge is gameable}
\label{app:diagnosis}

The probe uses an agentic game-builder (a
separate setting from the from-scratch generator studied in the main text) and shares the
benchmark's exact evaluation stack: the official Godot~4.6.2 runtime, headless
replay, and the official GPT-5.5 \textsc{Overall} visual judge (the same judge
used on the public leaderboard, so scores are directly comparable).

\paragraph{The agentic pipeline and the asset lever.}
The companion agent constructs a project through a sequence of build passes;
the one that matters here is a \emph{material-grounding} pass that
retrieves real CC0 sprite assets and substitutes them for the solid-color placeholder rectangles the agent otherwise emits. Enabling the material-grounding
pass raises the judge's art score with the game's code held fixed: the art score comes out $1.8\times$ what the same code earns with its original placeholders. The lever is presentation, not fit: \emph{randomly chosen} sprites of comparable provenance raise it as much as the retrieved ones. The manipulated
outputs move under a judge that is self-consistent on a fixed frame ($\sigma\!<\!0.01$). Across generation seeds, one model's \textsc{Overall} moves by $0.08$--$0.33$.

\section{Verification signals in detail}
\label{app:signals}

\begin{figure*}[t]
\centering
\includegraphics[width=\textwidth]{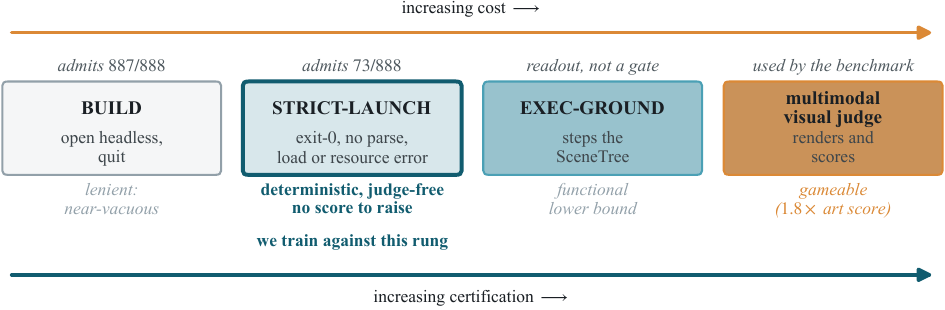}
\caption{\textbf{The verification ladder}, ordered by how much each signal certifies.
\textbf{\build{}} is near-vacuous (it admits $887$ of the supervised model's
$888$ training-family candidates, of which only $73$ also pass \slaunch{}); \textbf{\slaunch{}}, our training gate, requires a clean headless launch
(exit code $0$, no parse/load/resource error), deterministic and judge-free; \textbf{execution grounding} drives the running
SceneTree and serves as a readout rather than a gate; the top rung, a
multimodal visual judge, certifies the most but is gameable.}
\label{fig:ladder}
\end{figure*}

\paragraph{Why the lenient rung is near-vacuous.}
The lenient \build{} check (open the project headless and quit,
\texttt{godot -{}-headless -{}-quit-after}) does not treat parse, project, or
scene-resource errors as fatal: the process can still exit $0$. On the held-out
families, supervised-model outputs that pass this check are nonetheless
non-functional in concrete, diagnosable ways: scripts that fail to parse
(undeclared identifiers, calls to nonexistent methods) and so never load;
malformed \texttt{project.godot} files (unquoted values, bare input-map
identifiers) that cause the engine to open its project manager instead of the
game; and broken \texttt{.tscn} external-resource references. The same engine
renders the corresponding gold projects without error. Empirically this failure is
parse-dominated: of the $888$ candidates the supervised model generates on the training families, $815$ fail \slaunch{}, and every one of those contains a GDScript or \texttt{project.godot} parse error.

\paragraph{\slaunch{}.}
A candidate passes \slaunch{} if
\texttt{godot -{}-headless -{}-path PROJECT -{}-quit-after 5} returns exit code $0$
\emph{and} emits no parse, load, or resource-resolution error on \texttt{stderr} (we match patterns such as \texttt{Parse Error}, \texttt{Failed to load script}, unresolved identifiers or calls, and failed \texttt{ext\_resource} lookups). The gate separates working from broken projects:
all held-out gold projects pass \slaunch{} (the $25/25$ gold ceiling), and the \build{}-passing outputs that fail it do so with concrete parse or load errors.

\paragraph{Execution grounding (\textsc{exec-ground}).}
It injects an instrumentation probe into a temporary copy of the project and steps
the headless SceneTree for up to $180$ frames in passive, synthetic, and
demo-replay modes, replaying up to three recorded \texttt{demo\_outputs} traces
per task. The score is
$\mathbf{1}[\text{runtime-error-free}]\cdot(0.35\,s_{\text{state}}
+0.25\,s_{\text{passive}}+0.40\,s_{\text{active}})$, thresholded at $0.50$ for the grounded call. The state term reads script-exposed member variables (so single-node
procedural games still register state); the passive term measures state change with no input at all; the active term subtracts that baseline so autonomous animation is not counted as interaction. On the
calibration set ($15$ gold projects across families; $15$ \slaunch{}-broken
candidates plus one hand-built empty shell), gold scores a median of $0.75$
(minimum $0.58$) and every negative scores near zero (empty shell $0.013$): the threshold at $0.50$ separates the two sets with no overlap.

\paragraph{The brief-compliance negative control.}
As the negative reference for the brief-compliance measurement of
\S\ref{sec:results}, we wrote a single brief-agnostic project: one Node2D that
drifts, one that follows the pointer, and a bank of ticking member
variables. We submitted it verbatim for all $25$ held-out briefs. It launches cleanly, and the blind judge rates it $0/25$ against gold's $19/25$. The shell generator will be released with the code.

\section{Training, generation, and data}
\label{app:training}

\begin{figure}[tb]
\centering
\includegraphics[width=\columnwidth]{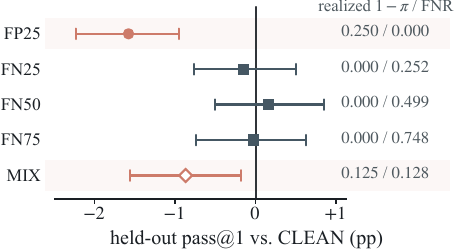}
\caption{\textbf{Only one direction of verifier error is paid for.} Held-out
pass@$1$ relative to the \textsc{clean} arm on APPS, in percentage points, with
$95\%$ crossed seed$\times$task bootstrap intervals, at a common count-matched
quota over eight training seeds. Realized, not nominal: $1\!-\!\pi$ is the
non-all-pass share of the \emph{admitted} rows (the axis of
Fig.~\ref{fig:dose}); the FNR is masked true positives over those available.
\textsc{fn} arms mask true positives,
\textsc{fp25} injects false ones, \textsc{mix} splits the error budget evenly.}
\label{fig:errdir}
\end{figure}

\begin{algorithm}[tb]
\caption{One round of launch-gated self-distillation.}
\label{alg:loop}
\SetKwInOut{Input}{input}\SetKwInOut{Output}{output}
\Input{model $M_{t}$; training-family briefs $\mathcal{B}$; gold pool
       $\mathcal{D}_{t}$; harvest budget $K_{\mathsf{h}}$: samples per brief}
\For{each brief $b \in \mathcal{B}$}{
  sample $K_{\mathsf{h}}$ candidate projects
  $\{c_1,\dots,c_{K_{\mathsf{h}}}\} \sim M_{t}(\cdot\mid b)$ \;
  \For{each candidate $c$ with at least three files}{
    materialize $c$ to a project tree \;
    \lIf{\slaunch{}$(c)$}{mark $c$ as clean}
  }
  keep up to two deduplicated clean candidates for $b$ \;
}
$\mathcal{F}_{t} \leftarrow$ all kept clean candidates (the round's \emph{fuel}) \;
$\mathcal{D}_{t+1} \leftarrow \mathcal{D}_{t} \cup \mathcal{F}_{t}$ \;
$M_{t+1} \leftarrow \textsc{fine-tune}(\text{base}, \mathcal{D}_{t+1})$ \;
\Output{$M_{t+1}$, $\mathcal{F}_{t}$}
\end{algorithm}

\begin{figure*}[t]
\centering
\includegraphics[width=\textwidth]{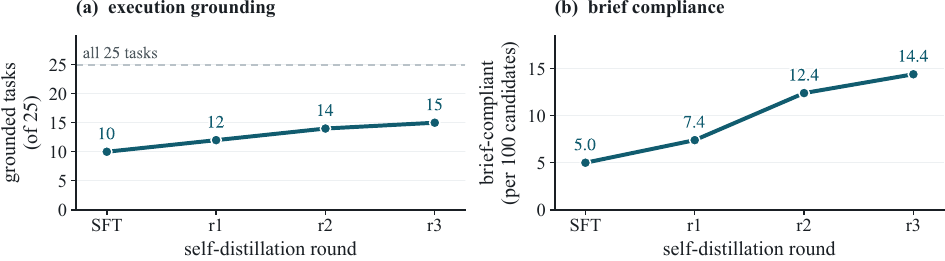}
\caption{\textbf{Two readouts the gate never optimizes.}
\textbf{(a)} Execution grounding: tasks with at least one grounded
($\ge\!0.50$) clean candidate, $10\!\to\!12\!\to\!14\!\to\!15$ of the same $25$
held-out tasks; the score is validated by $15/15$ grounded gold against $0/16$
known-bad projects. \textbf{(b)} Brief compliance under a blind judge that
reads only the brief and the shipped GDScript: projects per $100$ candidates
that both launch and implement their brief.}
\label{fig:ground}
\end{figure*}

\begin{figure}[tb]
\centering
\includegraphics[width=\columnwidth]{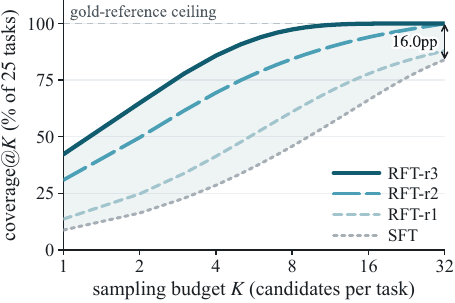}
\caption{\textbf{The supervised model never closes the coverage gap within the sampled budget.} Coverage@$K$
under the unbiased estimator $1-\binom{n-c}{K}\!/\!\binom{n}{K}$, pooling four
decoding seeds into $n\!=\!32$ candidates per task, $c$ of them clean, averaged
over the $25$ held-out tasks. The supervised model is still $16.0$pp short of
full coverage at $K\!=\!32$; round~3 has been above $97\%$ since $K\!=\!8$.}
\label{fig:covk}
\end{figure}

\begin{figure}[tb]
\centering
\includegraphics[width=\columnwidth]{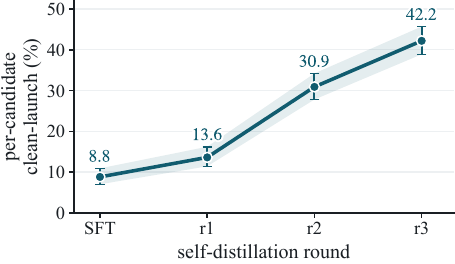}
\caption{\textbf{Three rounds compound.} The \pcand{} clean-launch rate on the four held-out
families climbs $8.8\!\to\!13.6\!\to\!30.9\!\to\!42.2\%$ with Wilson intervals.
Task-level coverage is not co-plotted; at the full
$32$-candidate budget it runs $84\!\to\!88\!\to\!100\!\to\!100\%$, saturating at
round~2. This series is the single-run trajectory the matched controls are
compared against; the three-seed means are in Table~\ref{tab:main}.}
\label{fig:compound}
\end{figure}

\begin{table}[tb]
\centering\small
\caption{\slaunch{} fuel rate per candidate, under a \emph{matched} training recipe ($8$ epochs), measured on the canonical $111$ training tasks ($888$ candidates per base).}
\label{tab:fivebase}
\setlength{\tabcolsep}{4pt}
\begin{tabular}{lrl}
\toprule
Base & \slaunch{} rate & Wilson $95\%$ CI \\
\midrule
Qwen3-4B          & $2.14\%$ & $[1.37,\ 3.32]$ \\
Llama-3.1-8B      & $2.93\%$ & $[2.01,\ 4.26]$ \\
Qwen3-8B          & $4.62\%$ & $[3.42,\ 6.20]$ \\
Qwen3-14B         & $8.22\%$ & $[6.59,\ 10.21]$ \\
\textbf{Gemma-3-12B} & $\mathbf{9.57\%}$ & $[7.81,\ 11.69]$ \\
\bottomrule
\end{tabular}
\end{table}

\begin{table}[tb]
\centering
\footnotesize
\setlength{\tabcolsep}{3pt}
\caption{Training-signal diagnostics for the GRPO arms over the same eight
training seeds as Fig.~\ref{fig:errdir}, as mean\,$\pm$\,s.d. over those seeds;
``clean reward'' scores the policy's own rollouts with the \emph{uncorrupted}
verifier.}
\label{tab:grpo}
\begin{tabular*}{\columnwidth}{@{\extracolsep{\fill}}lccc}
\toprule
 & \textsc{clean} & \textsc{fp25} & \textsc{fn25} \\
\midrule
degenerate groups & $0.488_{\pm.023}$ & $0.367_{\pm.062}$ & $0.141_{\pm.037}$ \\
gradient norm     & $0.293_{\pm.027}$ & $0.318_{\pm.034}$ & $0.405_{\pm.057}$ \\
clean reward      & $0.696_{\pm.036}$ & $0.629_{\pm.049}$ & $0.608_{\pm.028}$ \\
\bottomrule
\end{tabular*}
\end{table}

\paragraph{Separating precision from fuel quantity.}
The count-matched ladder varies precision by replacing admitted slots, so a
lower-precision arm also holds fewer true positives. A four-arm ladder at the
same $944$-row quota and the same eight training seeds, retrained and
re-evaluated end to end on a different accelerator, separates them:
\textsc{clean}; \textsc{fp25} (the same persisted fuel as
Fig.~\ref{fig:errdir}, $708$ true positives and $236$ false ones);
\textsc{c708} (those $708$ alone, at $708$ rows and so a quarter fewer
optimizer steps); and \textsc{c708r} (those same $708$, with $236$ of them
duplicated back to $944$ rows). Against \textsc{clean}: \textsc{c708r}
$+0.40$pp $[-0.18,+1.03]$, \textsc{c708} $-0.91$pp $[-1.86,+0.06]$, and
\textsc{fp25} $-2.17$pp $[-3.02,-1.46]$ with all eight seeds negative, which
reproduces the $-1.58$pp $[-2.23,-0.95]$ of the original run on overlapping
intervals. Both paths land on that penalty: the remaining step to \textsc{fp25} costs
$1.27$pp from \textsc{c708} and $2.57$pp from \textsc{c708r}. \textsc{c708} is handicapped on updates and still ahead of \textsc{fp25};
\textsc{c708r} cannot reproduce the per-problem slot histogram exactly, since a
replaced slot on a problem with no surviving sibling has nothing same-problem to
duplicate.

\paragraph{GRPO arms of the error-direction ladder.}
Each run trains Qwen3-4B with LoRA for $300$ optimizer steps at group size $G\!=\!8$ on $50$
problems whose pass rate under the current policy lies in $[0.25,0.75]$. The two corrupted arms
realize error rates of $24.97\%$ and $24.98\%$, giving Youden indices of $0.7503$ and $0.7502$.

\paragraph{Fuel expansion at 8B.}
The generator's own yield of \slaunch{}-clean candidates on the training
families covers $31\!\to\!51\!\to\!75$ of the $111$ tasks.

\paragraph{Data and split.}
The family partition is given in \S\ref{sec:setup}. We
verified zero held-out leakage into any round's fuel: every harvested brief
carries a training-family label and matches no held-out brief. Per round, the harvest
keeps, among candidates with at least three files, the \slaunch{}-clean ones, deduplicated,
capped at two per task: of $M_0$'s $73$ \slaunch{}-clean candidates this yields $67$
fuel projects, then $123$ from $M_1$ and $186$ from $M_2$; these cover $48$, $77$,
and $101$ of the $111$ training tasks across the three rounds, reaching all $10$
training families by round~2. Code, the harvested fuel, and one
adapter per reported arm will be released upon publication.

\paragraph{Training.}
On Qwen3-14B we train for $8$ epochs; hyper-parameters follow. All training and inference ran on a single $96$\,GB GPU; a full three-round loop (generation, filtering, and retraining) took on the order of tens of GPU-hours.

\paragraph{Generation.}
vLLM~0.11.2 serves the LoRA adapter unmerged at \texttt{bfloat16}. We draw $K_{\mathsf{h}}\!=\!8$ candidates for each training-family brief. Prompts are
rendered with each base model's own chat template
(\texttt{apply\_chat\_template}, \texttt{add\_generation\_prompt=True},
\texttt{enable\_thinking=False}), so each base is queried in its
native format. Decoding on Godot uses Qwen3 non-thinking sampling:
temperature $0.7$, top-$p$ $0.8$, top-$k$ $20$, repetition penalty $1.05$, and a
$24{,}576$-token generation cap. The cap is set above the longest gold reference
($19{,}192$ tokens).

\paragraph{LoRA and optimization.}
Adapters use $r\!=\!16$, $\alpha\!=\!32$, dropout $0.05$, no bias, task type
\textsc{causal-lm}. Target modules are resolved from
\texttt{named\_modules} at run time, matching the
\texttt{\{q,k,v,o,gate,up,down\}\_proj} projections; if a base carries a vision
tower, its projections are excluded and only language-model projections are
adapted. Optimization is AdamW with a cosine schedule and a $3\%$ warmup ratio,
learning rate $10^{-4}$, batch size $1$ with gradient accumulation $8$, sequence
length $24{,}576$, and scaled dot-product attention, all in \texttt{bfloat16}. The
prompt is masked out of the loss (labels set to $-100$ over the prompt span).
On APPS we reuse the same settings at the smaller scale: $r\!=\!16$, $\alpha\!=\!32$, learning rate $10^{-4}$, sequence length
$1536$; held-out generation uses temperature $0.8$ and top-$p$ $0.95$.

\paragraph{Evaluation and statistics.}
The 14B arms are evaluated with $8$ candidates per held-out task across four decoding seeds ($800$ candidates per model, pooled to $K\!=\!32$ for coverage); the 8B and 4B replications use five and six decoding seeds, respectively. The primary metric is
the \pcand{} clean-launch rate; comparisons between models are paired by task and
tested with a cluster-permutation test that permutes candidate labels within
each task ($N\!=\!25$ tasks, $50{,}000$ draws), with task-level bootstrap
confidence intervals on effect sizes ($50{,}000$ resamples). The APPS ladders
instead use a crossed seed$\times$task bootstrap: training seeds and held-out
problems are resampled independently with replacement and the effect is
recomputed over the crossed resamples ($10{,}000$ draws), with percentile
intervals. Because candidates drawn for the same task are strongly correlated (intraclass correlation $0.63$), every comparison clusters on the task rather than on the candidate. Wilson intervals are reported per arm as a description of its candidate pool and are not used for between-arm inference. Table~\ref{tab:protocol} lists the evaluation budget and
coverage estimator for every configuration.

\begin{table}[tb]
\centering
\small
\caption{Per-family coverage@$16$ at a single two-decoding-seed budget, in task
counts. Round~3 reaches full coverage in every family, while the
exact-duplication control (\ctrl{}) regresses in three of the four families.}
\label{tab:family}
\begin{tabular*}{\columnwidth}{@{\extracolsep{\fill}}l c c c c c}
\toprule
family  & SFT & r1 & r2 & r3 & \ctrl{} \\
\midrule
horror  & $5/5$ & $5/5$ & $5/5$ & $\mathbf{5/5}$ & $4/5$ \\
puzzle  & $5/8$ & $5/8$ & $7/8$ & $\mathbf{8/8}$ & $2/8$ \\
rhythm  & $2/5$ & $4/5$ & $4/5$ & $\mathbf{5/5}$ & $4/5$ \\
shooter & $6/7$ & $5/7$ & $6/7$ & $\mathbf{7/7}$ & $3/7$ \\
\bottomrule
\end{tabular*}
\end{table}

\begin{table}[tb]
\centering
\footnotesize
\setlength{\tabcolsep}{3pt}
\caption{Evaluation protocol per configuration. Draws pool all decoding seeds
($8$ candidates per task per seed; APPS rows are labeled by held-out size). Table~\ref{tab:family} instead uses one fixed seed pair, in task counts.}
\label{tab:protocol}
\begin{tabular*}{\columnwidth}{@{\extracolsep{\fill}}l c c l}
\toprule
configuration & seeds & draws & coverage \\
\midrule
Godot 14B & $4$ & $800$  & coverage@$32$, unbiased \\
Godot 8B  & $5$ & $1000$ & coverage@$32$, unbiased \\
Godot 4B  & $6$ & $1200$ & coverage@$48$, unbiased \\
APPS 4B /$150$ & $3$ & $3600$ & coverage@$8$, task counts \\
APPS 4B /$470$ & $3$ & $11280$ & coverage@$8$, task counts \\
\midrule
Fig.~\ref{fig:covk} (14B) & $4$ & $800$ & unbiased, $K\!\le\!32$ \\
\bottomrule
\end{tabular*}
\end{table}

\begin{table}[tb]
\centering
\footnotesize
\setlength{\tabcolsep}{2pt}
\caption{APPS-introductory, Qwen3-4B. From a lightly supervised $M_0$
($20.4\%$), only the \textsc{strict} (all-tests-pass) gate produces decisive transfer; the
count-matched \textsc{partial} ($\ge\!50\%$ but not all tests) control moves the rate less; \textsc{lenient} (syntax-valid only) leaves pass@$1$ untouched while cutting coverage@$8$. Paired bootstrap over
$470$ held-out problems, eight training seeds; rates in \%, $\Delta$ in percentage points, $p$ for the comparison with $M_0$. $^{*}\,p\!<\!0.05$, $^{**}\,p\!<\!0.01$.}
\label{tab:apps}
\begin{tabular*}{\columnwidth}{@{\extracolsep{\fill}}lcccc}
\toprule
Fuel gate & pass@$1$ & $\Delta$ & coverage@$8$ & $\Delta$ \\
\midrule
$M_0$ (gold SFT) & $20.4$            & ---          & $41.5$            & --- \\
\textsc{strict}  & $\mathbf{25.5}$   & $+5.15^{**}$ & $\mathbf{44.8}$   & $+3.32^{**}$ \\
\textsc{partial} & $22.5$            & $+2.11^{**}$ & $41.0$            & $-0.46$ \\
\textsc{lenient} & $20.3$            & $-0.10$      & $38.3$            & $-3.16^{**}$ \\
\bottomrule
\end{tabular*}
\end{table}

\section{Efficiency, not reach: what a zero at \texorpdfstring{$K\!=\!32$}{K=32} means}
\label{app:reach}

Four of the $25$ held-out tasks produced no clean project anywhere in the supervised model's full $32$-candidate pool, while the distilled model (RFT-r3) produced one on all four. We drew $512$ fresh candidates per task from the same frozen supervised
adapter, under the same gate and decoding configuration. All four yielded clean projects: $15$, $23$, $3$, and $5$ successes out of $512$, against $17$, $25$, $6$, and $8$ out of $32$ for the distilled
model. To reach, with $m$ supervised draws, the chance that a single distilled draw is clean ($1-(1-q)^{m}\!=\!q_{\textsc{rft}}$, where $q$ and $q_{\textsc{rft}}$ are the per-draw clean rates of the supervised and distilled models), the supervised model needs $25$--$35$ draws, so under a multi-file executable gate the
separation is one of sampling efficiency, placing this setting in the regime
\citet{yue2025rlvr} report for RLVR in the language domain. The reading survives a
larger budget: at 8B, \emph{one} launch-gated round covers $19/25$ at $K\!=\!40$, against $14/25$ for the supervised model and $15/25$ for one \build{}-gated round.

\end{document}